\documentclass[10pt,twocolumn,letterpaper]{article}

\usepackage{cvpr}              
\usepackage[accsupp]{axessibility}

\usepackage{graphicx}
\usepackage{amsmath}
\usepackage{amssymb}
\usepackage{booktabs}
\usepackage{multirow}
\usepackage{paralist} 
\let\itemize\compactitem 
\let\enditemize\endcompactitem 
\let\enumerate\compactenum 
\let\endenumerate\endcompactenum 
\let\description\compactdesc 
\let\enddescription\endcompactdesc

\usepackage[pagebackref,breaklinks,colorlinks]{hyperref}

\usepackage[capitalize]{cleveref}
\crefname{section}{Sec.}{Secs.}
\Crefname{section}{Section}{Sections}
\Crefname{table}{Table}{Tables}
\crefname{table}{Tab.}{Tabs.}

\def\cvprPaperID{10578} 
\def\confName{CVPR}
\def\confYear{2022}

\begin{document}

\makeatletter
\newcommand{\printfnsymbol}[1]{%
  \textsuperscript{\@fnsymbol{#1}}%
}
\makeatother

\title{TransVPR: Transformer-Based Place Recognition with \\ Multi-Level Attention Aggregation}

\author{Ruotong Wang\thanks{Equal contribution.} \quad Yanqing Shen\printfnsymbol{1} \quad Weiliang Zuo \quad Sanping Zhou \quad Nanning Zheng\thanks{Corresponding author.}\\
Institute of Artificial Intelligence and Robotics, Xi’an Jiaotong University\\
{\small \tt \{wrt072@stu., qing1159364090@stu., weiliang.zuo@, spzhou@, nnzheng@mail.\}xjtu.edu.cn}}

\maketitle

\begin{abstract}
   Visual place recognition is a challenging task for applications such as autonomous driving navigation and mobile robot localization. Distracting elements presenting in complex scenes often lead to deviations in the perception of visual place. To address this problem, it is crucial to integrate information from only task-relevant regions into image representations. In this paper, we introduce a novel holistic place recognition model, TransVPR, based on vision Transformers. It benefits from the desirable property of the self-attention operation in Transformers which can naturally aggregate task-relevant features. Attentions from multiple levels of the Transformer, which focus on different regions of interest, are further combined to generate a global image representation. In addition, the output tokens from Transformer layers filtered by the fused attention mask are considered as key-patch descriptors, which are used to perform spatial matching to re-rank the candidates retrieved by the global image features. The whole model allows end-to-end training with a single objective and image-level supervision. TransVPR achieves state-of-the-art performance on several real-world benchmarks while maintaining low computational time and storage requirements.
\end{abstract}
\section{Introduction}
\begin{figure}[t]
  \centering
   \includegraphics[width=1\linewidth]{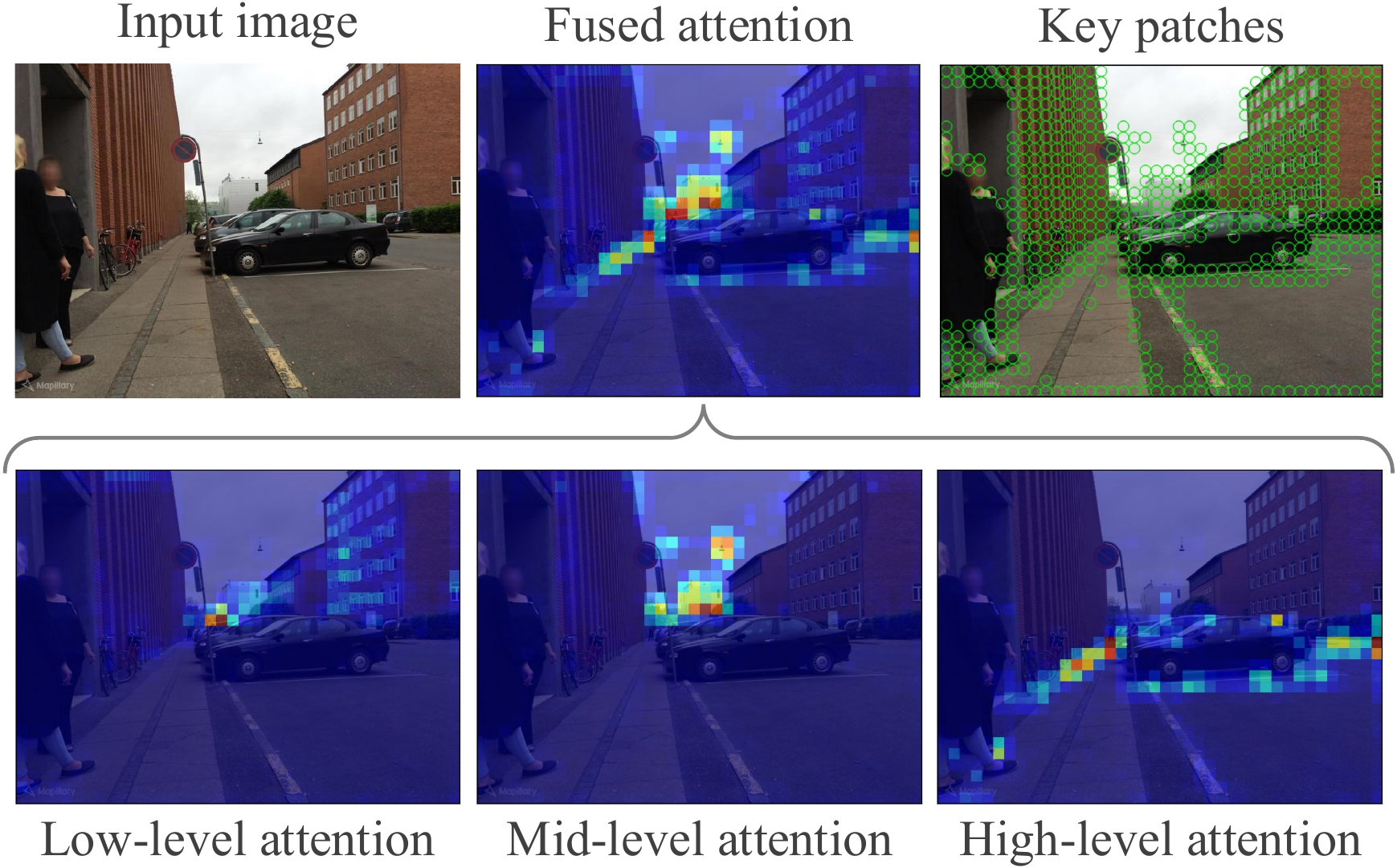}
   \caption{\textbf{Visualization of multi-level attentions from TransVPR.} Low-level attention maps mainly focus on small objects and textural areas on the surface of buildings. Mid-level attentions focus on objects in the air, such as street lamps and tree canopies, while high-level attentions tend to outline the contours of the ground and the lane lines. All these attention masks are combined to generate global image representations as well as key-patch descriptors.}
   \label{fig:show}
\end{figure}

Visual Place Recognition (VPR) is an essential and challenging problem in autonomous driving and robot localization systems, which is usually defined as an image retrieval problem\cite{lowry2015visual}. Given a query image, the algorithm has to determine whether it is taken from a place already seen and identify the corresponding images from a database. There are two types of image representations commonly used in VPR tasks. Global image features\cite{jegou2011aggregating,torii201524,arandjelovic2016netvlad,chen2017deep,gordo2017end, radenovic2018fine,revaud2019learning} abstract the whole image into a compact feature vector without geometrical information. Patch-level descriptors\cite{lowe1999object,yi2016lift,noh2017large, detone2018superpoint,dusmanu2019d2,khaliq2019holistic} describe particular patches or keypoints in an image and can be used to perform spatial matching between image pairs using cross-matching algorithms (\eg RANSAC\cite{fischler1981random}). To achieve a good trade-off between accuracy and efficiency, a commonly used two-stage strategy is to retrieve candidates with global features and then re-rank them using patch-level descriptor matching\cite{taira2018inloc, sarlin2019coarse}. Several recent researches\cite{schuster2019sdc,teichmann2019detect,cao2020unifying,hausler2021patch} have attempted to design a holistic system to extract both types of features. Most recently, Patch-NetVLAD\cite{hausler2021patch} has used an integral feature space to derive patch descriptors from the global image feature and has achieved state-of-the-art performance in several benchmarks. However, an important factor that may reduce the robustness of Patch-NetVLAD is that its extracted features unselectively encode the information from all regions of an image. 

It needs to be emphasized that the ability to identify task-relevant regions in an image is critical to VPR systems. This is because distracting elements and dynamic objects in a scene (\eg the sky, the ground, untextured walls, cars, pedestrians, \etc) are not helpful to recognize a place and seriously harm the VPR performance\cite{lowry2015visual}. In order to detect keypoints or regions of interest, several CNN-based methods have been proposed\cite{verdie2015tilde,yi2016lift, kim2017learned,detone2018superpoint, khaliq2019holistic, chen2018learning,yan2021hierarchical}.

Recently, the Transformer\cite{vaswani2017attention} architecture has obtained competitive results in multiple computer vision tasks\cite{dosovitskiy2020image,carion2020end}. Unlike CNNs, the self-attention operation in vision Transformers can dynamically aggregates global contextual information and implicitly select task-relevant information. To benefit from this property of vision Transformers and improve the robustness of place recognition, this work brings the following \textbf{contributions}: Firstly, we propose a Transformer-based novel place recognition model, TransVPR, which can adaptively extract robust image representations from distinctive regions in an image. Secondly, inspired by previous studies on CNNs which combine multi-level feature maps to enrich image representations\cite{yu2017multi,chen2018learning,yan2021hierarchical}, we fuse multi-level attentions, which focus on different semantically meaningful regions (see \cref{fig:show}), to generate global image representations. The effectiveness of this procedure is demonstrated by qualitative and quantitative experiments. Finally, the output tokens of Transformer layers filtered by the fused attention mask are further employed as patch-level descriptors to perform geometrical verification. All components in TransVPR are tightly coupled so that the whole model allows end-to-end optimization with a single training objective and only image-level supervision. Experimental results show that the proposed TransVPR achieves superior performance on VPR benchmark datasets with low computational time and memory requirements. It outperforms the state-of-the-art VPR approaches\cite{arandjelovic2016netvlad,ge2020self,cao2020unifying,sarlin2020superglue,hausler2021patch} by significant margins (5.8$\%$ absolute increase on Recall@1 compared with the best baseline method, DELG\cite{cao2020unifying}). 

\section{Related Work}
We review previous works on image description techniques, especially related to place recognition.

{\bf Patch-Level Descriptors.}  In early VPR systems\cite{se2002mobile,newman2005slam,kovsecka2005global,angeli2008fast,cummins2008fab,mur2017orb}, traditional methods such as SIFT\cite{lowe1999object}, SURF\cite{bay2006surf} and ORB\cite{rublee2011orb} have been widely used to represent a small patch centered around a detected keypoint. However, these handcrafted features cannot handle severe appearance changes. More recently, CNN-based methods have achieved superior performances\cite{noh2017large, khaliq2019holistic, teichmann2019detect,camara2020visual,hausler2021patch}. In order to extract sparse patch descriptors, some approaches have proposed to firstly detect keypoints based on local structures and then describe them with a separate CNN\cite{yi2016lift,ono2018lf,savinov2017quad,zhang2018learning, simo2015discriminative}, while others have used shared network to preform both detection and description\cite{detone2018superpoint, dusmanu2019d2}. In addition to these general methods, several attempts have also been devoted to learn task specific patch-level features for place recognition\cite{noh2017large,cao2020unifying,yuan2021softmp}. Besides, Patch-NetVLAD\cite{hausler2021patch} provided an alternative solution using a global descriptor technique, NetVLAD\cite{arandjelovic2016netvlad}, to extract descriptors from pre-defined image patches. 

In most previous studies, patch-level descriptors refer to as \textit{local descriptors}, which encode the content in local patches around keypoints. In contrast, our Transformer-based patch-level descriptors are not local, since each output token from a Transformer layer has global perception field. In this way, the patch descriptors are able to capture more semantically meaningful structures with long-range dependency.

{\bf Global Image Representation.} Global image features are usually obtained by aggregating local descriptors. Some traditional techniques, such as Bag of Words (BoW)\cite{sivic2003video, csurka2004visual}, Fisher Kernel\cite{perronnin2007fisher,perronnin2010large, jegou2011aggregating}, and Vector of Locally Aggregated Descriptors (VLAD)\cite{jegou2010aggregating,arandjelovic2013all}, have been used to assign visual words to images. Likewise, in deep learning context, some works\cite{arandjelovic2016netvlad,mohedano2016bags} have incorporated these clustering methods into CNN architectures, while other studies\cite{tolias2016particular,gordo2017end,kalantidis2016cross,radenovic2018fine} have focused on pooling from CNN feature maps. Recently, unified networks have been developed to jointly extract global features and patch-level descriptors\cite{taira2018inloc,sarlin2019coarse,simeoni2019local,cao2020unifying}. Previous CNN-based approaches which extract features from high-level convolutional layers require deep networks with down sampling layers to integrate enough contextual information. As a first attempt, El-Nouby \etal\cite{el2021training} introduced vision Transformers to image retrieval tasks by using the {\tt [class]} token\cite{vaswani2017attention} from the final layer as a global feature. Differently, we aggregate multi-level attentions to generate global features and explicitly learn corresponding attention maps which can be further employed to detect key-patches.

{\bf Attentions for Place Recognition.}
In order to adaptively identify task-relevant regions in a complex scene image, attention mechanism has been adopted in several VPR approaches. Among them, the learned attention maps can be considered as patch descriptor filters\cite{noh2017large,yan2021hierarchical,yuan2021softmp} or weight maps which modulate the CNN feature maps to generate global features\cite{kim2017learned, chen2018learning}. The attention module in CNN-based methods has usually been implemented as a shallow CNN which is trained separately\cite{noh2017large} or jointly\cite{cao2020unifying,yuan2021softmp,chen2018learning,yan2021hierarchical} with the backbone network. In our work, a new formulation is proposed, where the attention module is as simple as a linear layer which decodes the attention information from Transformer tokens.

\begin{figure}[t]
  \centering
   \includegraphics[width=1\linewidth]{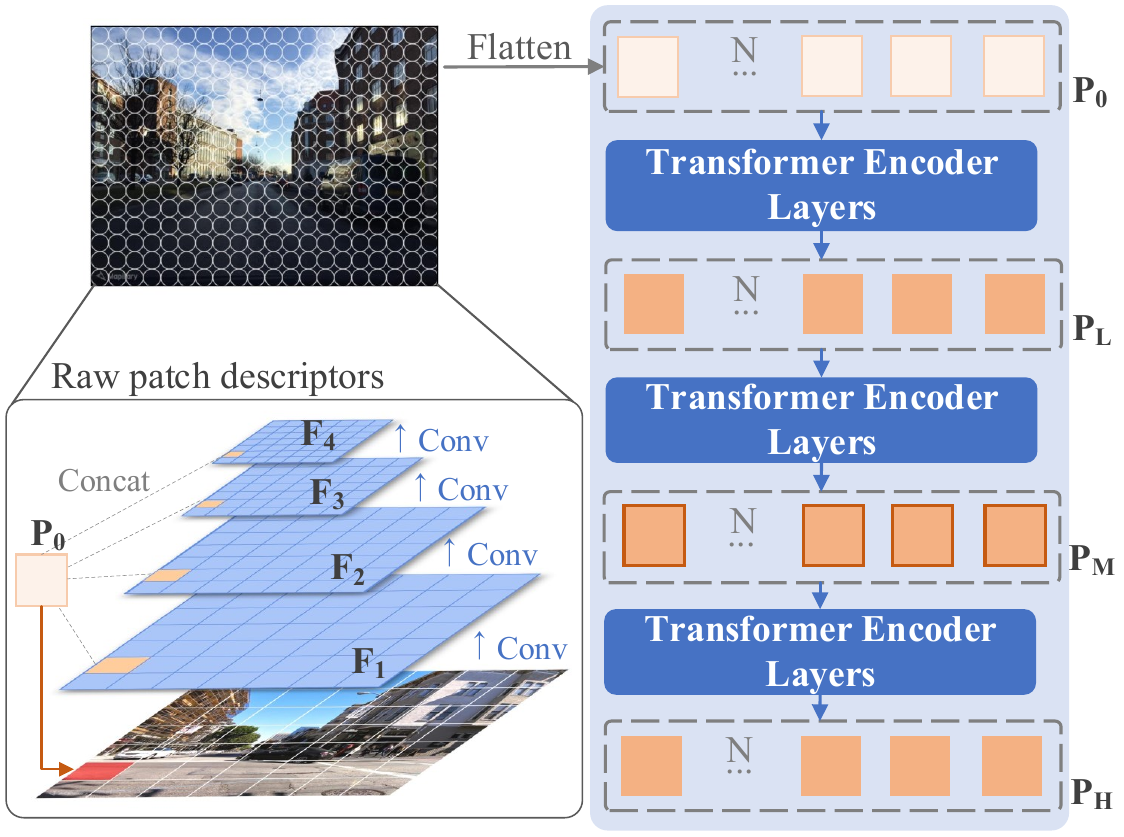}
   \caption{{\bf Patch descriptor extraction.} For an input image, pyramid feature maps are generated by a CNN and each reshaped into a sequence of flattened 2D patches. Raw patch-level descriptors are obtained by concatenating patch embeddings at the same position from each feature map. They are then sent into a Transformer encoder to integrate global contextual information. Output patch tokens from the low-level, mid-level, and high-level Transformer layers are selected for subsequent processing.}
   \label{fig:extraction}
\end{figure}

\section{Methodology}
 TransVPR jointly extracts both patch-level and global image representations by aggregating multi-level attentions in vision Transformers. Given an input image, its raw patch descriptors are firstly extracted by a shallow CNN and embedded as input tokens of a vision Transformer. Attentions from shallow, middle, and deep layers of the Transformer are merged to generate global image features and detect task-relevant patches. \cref{fig:extraction} and \cref{fig:aggregation} illustrate the whole feature extraction pipeline.

\subsection{Patch Descriptor Extraction}
A four-layer CNN is applied on an input image to extract raw patch-level features, as illustrated in \cref{fig:extraction}. Given an image or a feature map $\mathbf{F_{i-1}} \in \mathbb{R}^{H_{i-1} \times W_{i-1} \times C_{i-1}}$, the output of a convolutional layer is:
\begin{equation}
    \small
    \mathbf{F_i} = \mathrm{MaxPool(ReLU(BN(Conv(}\mathbf{F_{i-1}}\mathrm{))))},
    \label{eq:conv_layer}
\end{equation}
where $\mathbf{F_i} \in \mathbb{R}^{\frac{H_{i-1}}{2} \times \frac{W_{i-1}}{2} \times C_i}$. In practice, $3 \times 3$ convolutional kernel is used, and the number of output channels are set to 64, 128, 256, and 512 respectively. In this way, we obtain a feature pyramid $\{ \mathbf{F_1}, \mathbf{F_2}, \mathbf{F_3}, \mathbf{F_4} \}$, where the size of feature map is reduced by half in order.

Then, \textit{patch embedding} is applied on each feature map following the process proposed in \cite{dosovitskiy2020image}. A feature map $\mathbf{F_{i}} \in \mathbb{R}^{H_i \times W_i \times C_i}$ is reshaped into a sequence of flattened 2D patches $\mathbf{F'_{i}} \in \mathbb{R}^{N \times (R_i^2 \cdot C_i)}$, where $(R_i, R_i)$ is the resolution of a feature map patch and $N = H_i W_i/R_i^2$ is the number of patches. To maintain a fixed number of patches on each feature map, $R_i$ is set to $R_{i-1}/2$. The flattened patches are mapped to $D/4$ dimensions, where $D$ is the latent embedding dimension of the subsequent Transformer blocks.  

Next, patch embeddings at the same position of different feature maps are concatenated as a raw patch-level local descriptor. We denote the group of raw patch descriptors as $\mathbf{P_0} \in \mathbb{R}^{N \times D}$. The location of a patch descriptor is approximated to the center coordinate of the corresponding image patch.

Finally, in order to integrate global contextual information, raw patch descriptors are then sent into a Transformer encoder as input tokens. We follow the standard implementation of Transformer encoder, which consists of a stack of Multi-headed Self-Attention (MSA) and Multi-Layer Perceptron (MLP) modules\cite{vaswani2017attention, dosovitskiy2020image}. In pre-training, a learnable {\tt [class]} token is added in the front of the token sequence to obtain an image representation for classification. Since the spatial positional information can be implicitly encoded in raw patch descriptors by the CNN architecture, \textit{positional embedding} is removed from Transformer blocks so that the model can be flexible to different input sizes. 

\subsection{Multi-Level Attentions}
Although the Transformer has global perception field from the lowest layers, it is observed that its mean attention distance increases with depth\cite{dosovitskiy2020image}. In other words, there are some differences in the scales of structures captured by different Transformer layers. To integrate information across multiple levels, three groups of output patch tokens from the low-level, mid-level, and high-level layers of Transformer are selected, denoted as $\{ \mathbf{P_L}, \mathbf{P_M}, \mathbf{P_H} \}$. A group of multi-level patch tokens $ \mathbf{P}$ are firstly composed by concatenating these three groups of tokens along the channel.
\begin{equation}
    \small
    \mathbf{P} = \mathrm{Concat}([\mathbf{P_L}, \mathbf{P_M}, \mathbf{P_H}]) \in \mathbb{R}^{N \times 3D}.
    \label{eq:concat}
\end{equation}

\begin{figure}[t]
  \centering
   \includegraphics[width=1.02\linewidth]{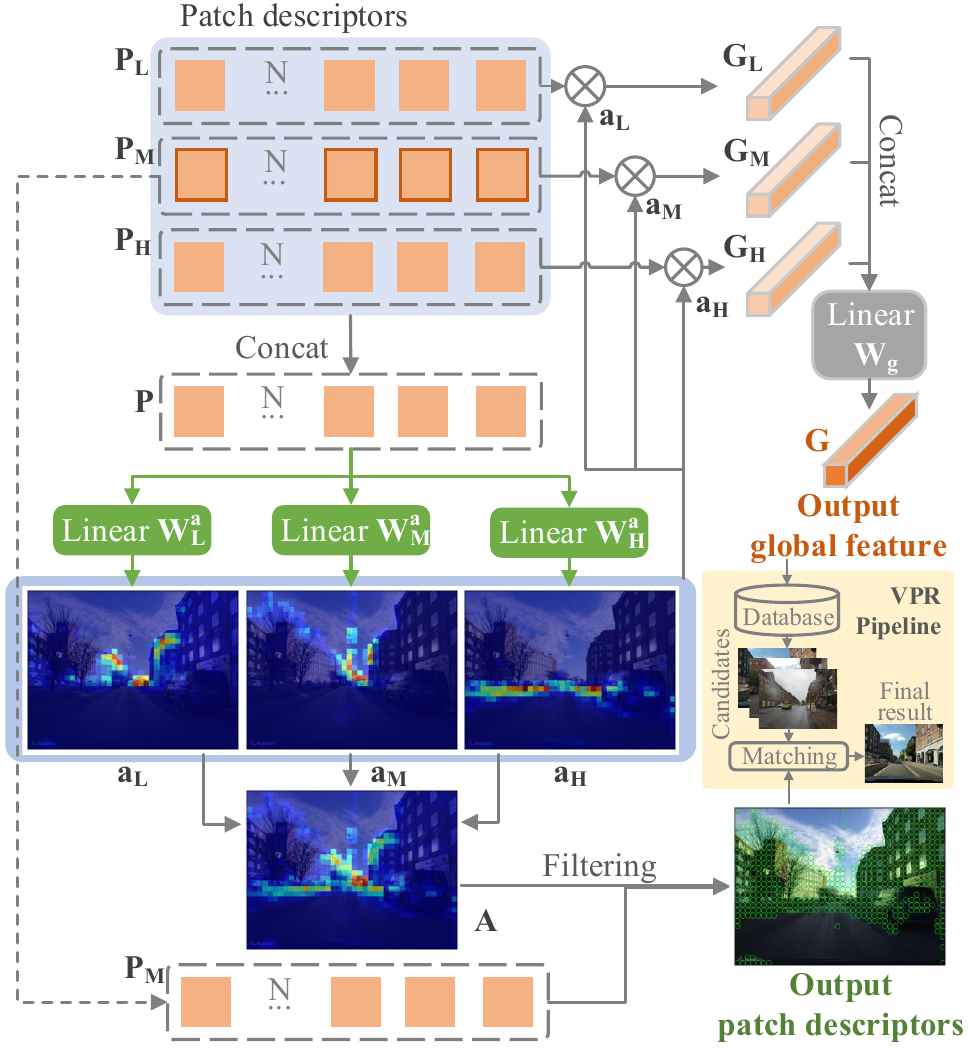}
   \caption{\textbf{Multi-level attention aggregation.} Given patch tokens from three different layers of Transformer, three attention maps are generated by applying a linear projection on their concatenation. The multi-level global feature is generated by combining single level global features, which are computed by summarizing the patch tokens weighted by the corresponding attention maps. Multi-level attention maps are further fused and used to select task-relevant patch descriptors. In VPR pipeline, global representations are used to retrieve candidates by nearest neighbour searching, while patch descriptors are used to perform geometrical verification to re-rank these candidates.}
   \label{fig:aggregation}
\end{figure}

For each level, an attention mask is estimated individually over all spatial positions, indicating the contribution of the information encoded in each specific patch token to recognize a place. Note that while computing these attention masks, the concatenated patch tokens $ \mathbf{P}$ are used (see \cref{fig:aggregation}). Formally:
\begin{equation}
    \small
    \mathbf{a_i} = \mathrm{softmax}(\mathbf{PW_i^a}) \in \mathbb{R}^{N \times 1},
    \label{eq:attention}
\end{equation}
where $i \in \{L,M,H\}$ and $\mathbf{W_i^a} \in \mathbb{R}^{3D \times 1}$ maps a concatenated patch token to a scalar. Then, a multi-level attention map $\mathbf{A} \in \mathbb{R}^{N \times 1}$ is generated by merging the three attention maps.
\begin{equation}
    \small
    \mathbf{A} = \mathrm{MinMaxNorm}(\sum_{i} \mathrm{MinMaxNorm}(\mathbf{a_i})).
    \label{eq:attention_fusion}
\end{equation}

\subsection{Final Image Representations}
\textbf{Key-Patch Descriptors.} In theory, output tokens from any Transformer layer can be used as patch-level descriptors to perform geometrical verification. In practice, we choose the mid-level patch tokens ($\mathbf{P_M}$) which give the most stable result in experiments. Patches where attention score $\mathbf{A}$ greater than a threshold $\tau$ are defined as key-patches, and their corresponding descriptors are used in final geometrical verification stage. 

\textbf{Global Image Features.} As illustrated in \cref{fig:aggregation}, global feature in a single level $\mathbf{G_i}$ is computed by aggregating patch tokens $\mathbf{P_i}$ which are weighted by the corresponding attention map $\mathbf{a_i}$:
\begin{equation}
    \small
    \mathbf{G_i} = \mathbf{a_i}^T\mathbf{P_i} \in \mathbb{R}^{D}.
    \label{eq:global}
\end{equation}

Multi-level global image feature $ \mathbf{G^*} \in \mathbb{R}^{3D}$ is defined as a concatenation of $\mathbf{G_L}$, $\mathbf{G_M}$ and $\mathbf{G_H}$, and then final global representation $\mathbf{G}$ is obtained by post-processing $\mathbf{G^*}$:
\begin{equation}
    \small
    \mathbf{G} = \mathrm{L2Norm(L2Norm}(\mathbf{G^*)W_g}),
    \label{eq:global}
\end{equation}
where learnable matrix $\mathbf{W_g} \in \mathbb{R}^{3D \times D}$ is used for dimensionality reduction.

\subsection{Training Strategy}
At first, the feature extraction backbone (CNN and Transformer) is pre-trained jointly on Places365\cite{zhou2017places}, an image classification dataset containing 1.8 million images from 365 scene categories. The {\tt [class]} token from the last Transformer layer is followed by a fully connected layer for classification.

Then, the pre-trained model is transferred to image retrieval task by removing the classification layer and adding the attention and the dimensionality reduction modules. The commonly used triplet margin loss\cite{schroff2015facenet} is adopted to be the training objective, defined as:
\begin{equation}
    \footnotesize
    L(\mathbf{G^q},\mathbf{G^p},\mathbf{G^n}) = \mathrm{max}(d(\mathbf{G^q},\mathbf{G^p})-d(\mathbf{G^q},\mathbf{G^n})+m,0), 
    \label{eq:triplet_loss}
\end{equation}
where $\mathbf{G^q}$, $\mathbf{G^p}$ and $\mathbf{G^n}$ are global features of query, positive and negative samples. The margin $m$ is a constant hyper-parameter. Parameters in attention and dimensionality reduction modules ($W_i^a$ and $W_g$) are initialized by training for a few epochs on a large scale VPR dataset, Mapillary Street Level Sequences (MSLS)\cite{warburg2020mapillary} training set, with frozen backbone parameters. 

After initialization, the whole TransVPR can be further finetuned in an end-to-end fashion on VPR datasets. 

\section{Experiments}
In this section, we evaluate the proposed TransVPR model on several benchmark datasets compared with some state-of-the-art VPR methods. We give the details of experimental settings, datasets, evaluation metrics, and compared methods in the following. 

\subsection{Implementation Details}
\textbf{Model Settings.} TransVPR is implemented in PyTorch framework. The base TransVPR model contains six transformer encoder layers for feature aggregation. The latent embedding dimension $D$ of the Transformer is 256. Without loss of generality, output tokens from the second, forth and sixth transformer layers are selected as $\mathbf{P_L}$, $\mathbf{P_M}$ and $\mathbf{P_H} $. The total parameter size of TransVPR is 19.86MB. The key-patch filtering threshold $\tau$ is set to 0.02 in practice. The patch size on the original image is set to $16 \times 16$. The dimensionality of output patch-level and global features are all set to 256.

In geometrical verification, given an image pair, their key-patch descriptors are matched in a brute-force manner. Cross checking is performed to ensure that matched descriptors are mutual nearest neighbors. The image similarity is defined as the number of inliers when estimating the homography based on the matched patches with RANSAC algorithm. Maximum allowed reprojection error of inliers is set to 1.5 times of the patch size. 

\textbf{Training.} We finetuned the pre-trained TransVPR model on MSLS training set and pittsburgh 30k (Pitts30k)\cite{torii2013visual} training set. The former aims to deal with evaluation datasets containing diverse scenes (MSLS and Nordland\cite{sunderhauf2013we,olid2018single} datasets) while the latter is particularly for urban scenes (Pitts30k and Robotcar Seasons V2\cite{Sattler2018CVPR, Maddern2017IJRR} datasets). In MSLS training set, both GPS coordinates and compass angles are provided, so the positive sample is selected as the image with the most similar field of view to the query. For Pitts30k dataset where the angle labels are not given, the weakly supervised positive mining strategy proposed in \cite{arandjelovic2016netvlad} is adopted. Hyper-parameters and further details for training are presented in Supplementary Material.

\subsection{Datasets}
We evaluated our model on several public benchmark datasets: MSLS\cite{warburg2020mapillary}, Nordland\cite{sunderhauf2013we,olid2018single}, Pitts30k\cite{torii2013visual} and RobotCar Seasons v2 (RobotCar-S2)\cite{Sattler2018CVPR, Maddern2017IJRR}. All of these datasets contain some challenging environmental variations. \cref{tab:datasets} summarizes the qualitative nature of them. More details of dataset usage are given in Supplementary Material. All images are resized to $640 \times 480$ while evaluation.

\begin{table}
  \centering
  \scalebox{0.76}{
  \begin{tabular}{@{}l|ccc|ccccc@{}}
  \toprule
  \multirow{2}{*}{Dataset} & \multicolumn{3}{c|}{Environment}  & \multicolumn{5}{c}{Variation}\\
  \cline{2-9}
    & \rotatebox{90}{Urban} & \rotatebox{90}{Suburban}& \rotatebox{90}{Natural} &\rotatebox{90}{Viewpoint } & \rotatebox{90}{Day/night } & \rotatebox{90}{Weather} & \rotatebox{90}{Seasonal} & \rotatebox{90}{Dynamic} \\
  \midrule
    MSLS\cite{warburg2020mapillary} & \checkmark & \checkmark &\checkmark&  $+$  & $+$  & $+$  & $+$  & $+$ \\
    Nordland\cite{sunderhauf2013we, olid2018single} & & \checkmark&\checkmark& $-$  & $-$  & $-$   & $+$  & $-$ \\
    Pitts30k\cite{torii2013visual} &\checkmark &&& $+$  & $-$  & $-$ & $-$  & $+$\\
    RobotCar-S2\cite{Sattler2018CVPR, Maddern2017IJRR}  & \checkmark &&& $+$  & $+$  & $+$  & $+$  & $+$\\
  \bottomrule
  \end{tabular}}
  \caption{\textbf{Summary of datasets used for evaluation.} $+$ indicates that the dataset contains the particular environmental variation, and $-$ is the opposite.}
  \label{tab:datasets}
\end{table}

\subsection{Metrics}
For MSLS, Nordland and Pitts30k datasets, we use Recall@N metric which computes the percentage of query images that are correctly localized. A query image is considered as correctly localized if at least one of the top $N$ ranked reference images is within a threshold distance from the ground truth location of the query. Default threshold definitions are used for all datasets\cite{warburg2020mapillary,olid2018single,torii2013visual}.

For RobotCar-S2 dataset, we follow \cite{hausler2021patch} and directly use the pose of the best matched reference image as the estimated pose of the query without calculating explicit 6-DOF poses. Recall@1 scores under three default error tolerances are used as evaluation metrics.

\begin{figure}[t]
  \centering
   \includegraphics[width=1\linewidth]{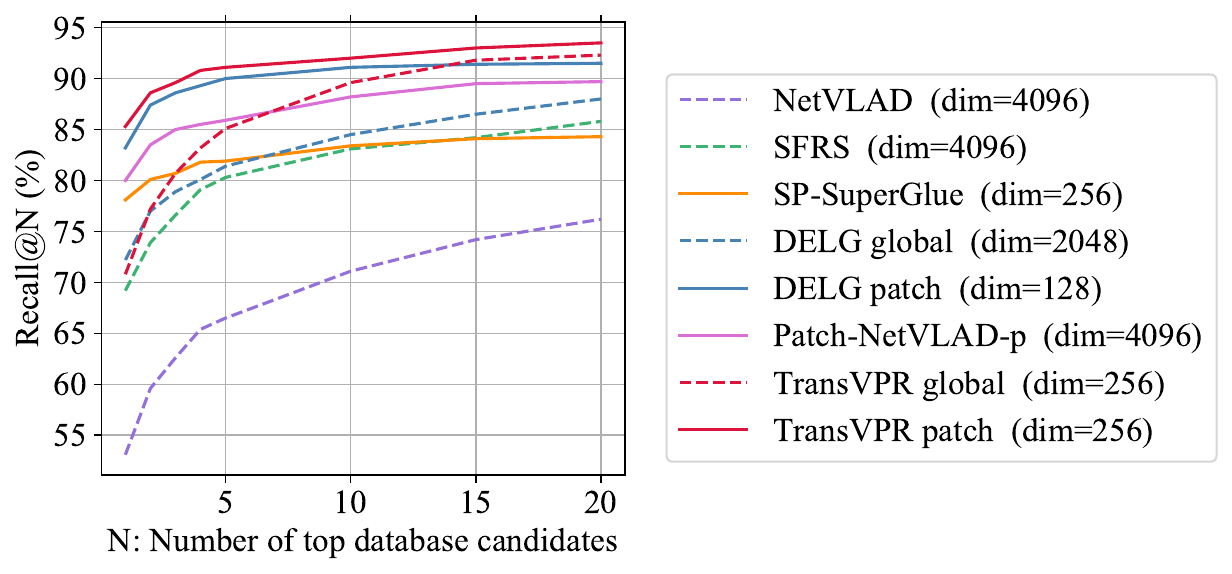}
   \caption{\textbf{Recall@N curves on MSLS val dataset.} Retrieval results using only global representation are depicted in dotted line, while results after re-ranking are depicted in solid line. TransVPR achieves the best performance in both global retrieval stage and re-ranking stage.}
   \label{fig:msls_compare}
\end{figure}

\begin{table*}
  \centering
  \scalebox{0.73}{
  \renewcommand{\arraystretch}{1.2}
  \begin{tabular}{@{}l||ccc||ccc||ccc||ccc||ccc@{}}
  \toprule
\multirow{2}{*}{Method}   & \multicolumn{3}{c||}{MSLS val} & \multicolumn{3}{c||}{MSLS challenge} & \multicolumn{3}{c||}{Nordland test} & \multicolumn{3}{c||}{Pitts30k test} & \multicolumn{3}{c}{Robotcar-S2 test}\\
\cline{2-16}
& R@1 & R@5 & R@10 & R@1 & R@5 & R@10 & R@1 & R@5 & R@10  & R@1 & R@5 & R@10  & .25m/2° & .5m/5° & 5.0m/10° \\
\hline
NetVLAD\cite{arandjelovic2016netvlad} & 53.1 & 66.5 & 71.1 &35.1 & 47.4 & 51.7 & 7.7 & 13.7 & 17.7 & 81.9 & 91.2 & 93.7 & 5.6 & 20.7 & 71.8 \\
SFRS\cite{ge2020self} & 69.2 & 80.3 & 83.1 & 41.5 & 52 & 56.3 & 18.8 & 32.8 & 39.8 & 89.4 & 94.7 & 95.9 & 8.0 & 27.3 & 80.4\\
TransVPR (w/o re-ranking) & 70.8 & 85.1 & 89.6  & 48.0 & 67.1 & 73.6 & 15.9 & 38.6 & 49.4 & 73.8 & 88.1 & 91.9 &2.9 & 11.4 & 58.6\\
\hline
SP-SuperGlue\cite{detone2018superpoint,sarlin2020superglue} & 78.1 & 81.9 & 84.3 & 50.6 & 56.9 & 58.3 &  29.1 & 33.5 & 34.3 & 87.2 & 94.8 & 96.4 & 9.5 & \textbf{35.4} & 85.4  \\
DELG\cite{cao2020unifying} & 83.2 & 90.0 & 91.1 & 52.2 & 61.9 & 65.4 & 51.3 & 66.8 & 69.8 & \textbf{89.9} & \textbf{95.4} & \textbf{96.7} & 2.2 & 8.4 & 76.8 \\
Patch-NetVLAD-s\cite{hausler2021patch} & 77.8 & 84.3 & 86.5 & 48.1 & 59.4 & 62.3 & 34.9 & 49.8 & 53.3 & 87.5 & 94.5 & 96.0 &  2.7 & 8.9 & 33.9\\
Patch-NetVLAD-p\cite{hausler2021patch} & 79.5 & 86.2 & 87.7 & 48.1 & 57.6 & 60.5 & 46.4 & 58.0 & 60.4 & 88.7 & 94.5 & 95.9  & 9.6 & 35.3 & \textbf{90.9} \\
\hline
TransVPR & \textbf{86.8} & \textbf{91.2} & \textbf{92.4} & \textbf{63.9} & \textbf{74.0} & \textbf{77.5} & \textbf{58.8} & \textbf{75.0} & \textbf{78.7} & 89.0 & 94.9 & 96.2 &\textbf{9.8} & 34.7 & 80.0\\
\bottomrule
\end{tabular}}
  \caption{Comparison to state-of-the-art methods on benchmark datasets.}
  \label{tab:compare_SOTA}
\end{table*}

\subsection{Compared Methods}
We compared TransVPR against several state-of-the-art algorithms, including two VPR methods based on nearest-neighbor searching using global image representations: \textbf{NetVLAD}\cite{arandjelovic2016netvlad} and \textbf{SFRS}\cite{ge2020self}, and two models which extract both global and patch descriptors for two-stage pipeline (\ie, retrieval and re-ranking): \textbf{Patch-NetVLAD}\cite{hausler2021patch} and \textbf{DELG}\cite{cao2020unifying}. For Patch-NetVLAD, we tested both its speed-focused and performance-focused configurations, denoted as Patch-NetVLAD-s and Patch-NetVLAD-p respectively. In addition, we also compared against a strong hybrid baseline, \textbf{SP-SuperGlue}, which re-ranks NetVLAD retrieved candidates by using SuperGlue\cite{sarlin2020superglue} matcher to match SuperPoint\cite{detone2018superpoint} patch-level descriptors. For all two stage methods, top-100 images retrieved by global features are further re-ranked by geometrical verification results. More installation details of the compared methods are explained in the Supplementary Material.
\begin{figure*}[t]
  \centering
   \includegraphics[width=0.98\linewidth]{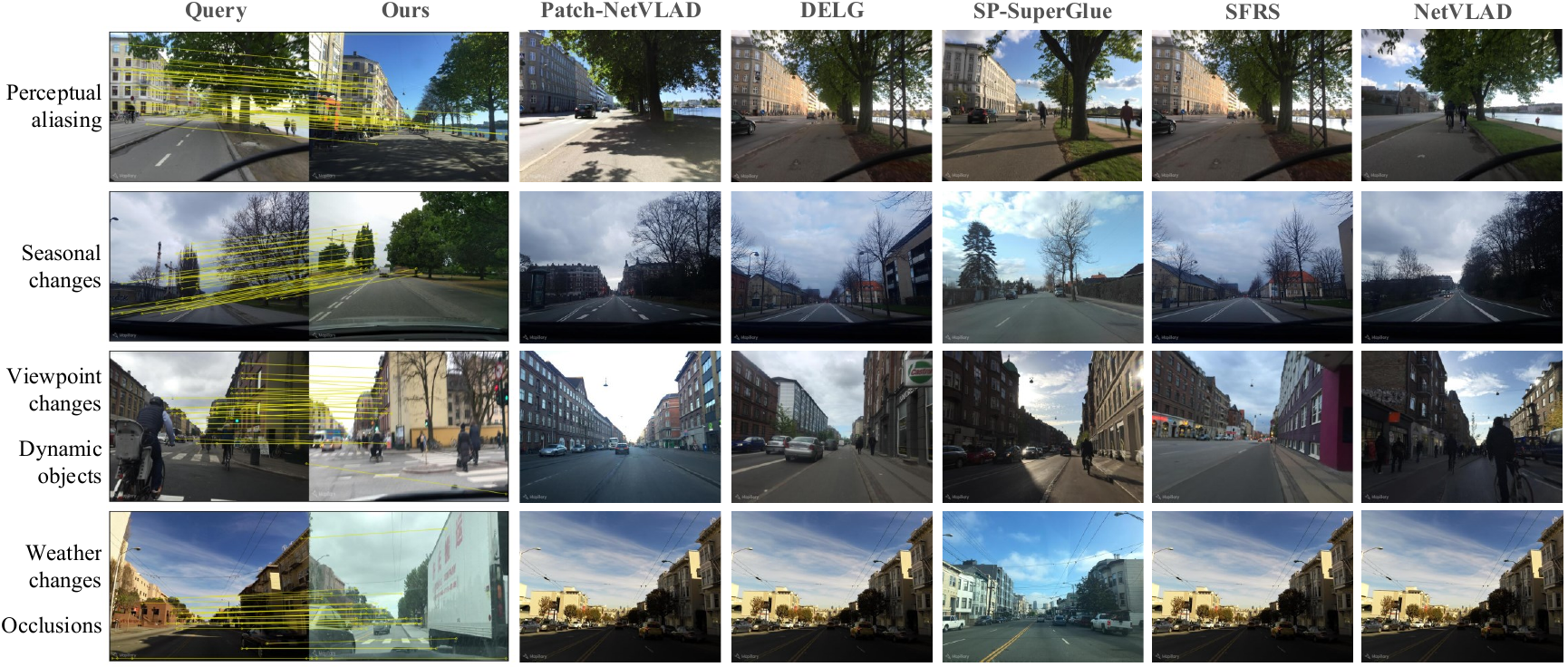}
   \caption{\textbf{Comparison of retrieval results on MSLS validation dataset.} In these challenging examples, TransVPR successfully retrieves the matching database image, while all other methods produce false results.}
   \label{fig:vis_compare}
\end{figure*}
\begin{table}
  \centering
  \scalebox{0.75}{
\begin{tabular}{l|c|c|c}
\toprule
Method & \begin{tabular}[c]{@{}c@{}}Extraction\\ latency (ms)\end{tabular} &         \begin{tabular}[c]{@{}c@{}}Matching\\ latency (s)\end{tabular} & \begin{tabular}[c]{@{}c@{}}Memory \\ (MB)\end{tabular} \\
\midrule
NetVLAD\cite{arandjelovic2016netvlad} & 17 & $-$ & $-$ \\
SFRS\cite{ge2020self} & 203 & $-$ & $-$ \\
SP-SuperGlue\cite{detone2018superpoint,sarlin2020superglue} & 166 & 7.83 & 1.93 \\
DELG\cite{cao2020unifying} & 197 & 36.04 & 0.37 \\
Patch-NetVLAD-s\cite{hausler2021patch} & 63 & 1.73 & 1.82 \\
Patch-NetVLAD-p\cite{hausler2021patch} & 1336 & 7.65 & 44.14 \\
\midrule
TransVPR (ours) & 45 & 3.19 & 1.17 \\
\bottomrule
\end{tabular}}
\caption{Feature extraction time, descriptor matching time, and memory footprint of all methods. Latency is measured on an NVIDIA GeForce RTX 2080 Ti GPU. For global retrieval methods, matching latency and memory requirements are negligible.}
\label{tab:latency_memory}
\end{table}
\section{Results and Discussion}
\subsection{Quantitative Results}
The quantitative results of TransVPR compared with other approaches are shown in \cref{tab:compare_SOTA}. Our TransVPR convincingly outperforms all compared methods on MSLS validation, MSLS challenge and Nordland datasets, with an absolute increase on Recall@1 of 3.6$\%$, 11.7$\%$ and 7.5$\%$ respectively in compared with the best baseline, DELG. It also achieves competitive results on Pitts30k and Robotcar-S2 datasets. Note that when training TransVPR on Pitts30k dataset, we only used the weakly supervised learning strategy in \cite{arandjelovic2016netvlad}, and we can expect further boost of TransVPR performance using the fine-grained supervision proposed by \cite{ge2020self}. Taking the average of all datasets, our full model surpasses the global feature retrieval based methods by a large margin, and outperforms the two-stage approaches, SP-SuperGlue, DELG and Patch-NetVLAD, with absolute gains of 10.8$\%$, 5.9$\%$ and 7.2$\%$ on Recall@1 score. 

The Recall@N curves of all methods including global retrieval results and re-ranking results are plotted in \cref{fig:msls_compare}. TransVPR also achieves the best result in global retrieval stage. Note that among all compared methods, TransVPR is the only method which selectively integrates task-relevant information when generating global representations. 

\begin{figure*}[t]
  \centering
   \includegraphics[width=.95\linewidth]{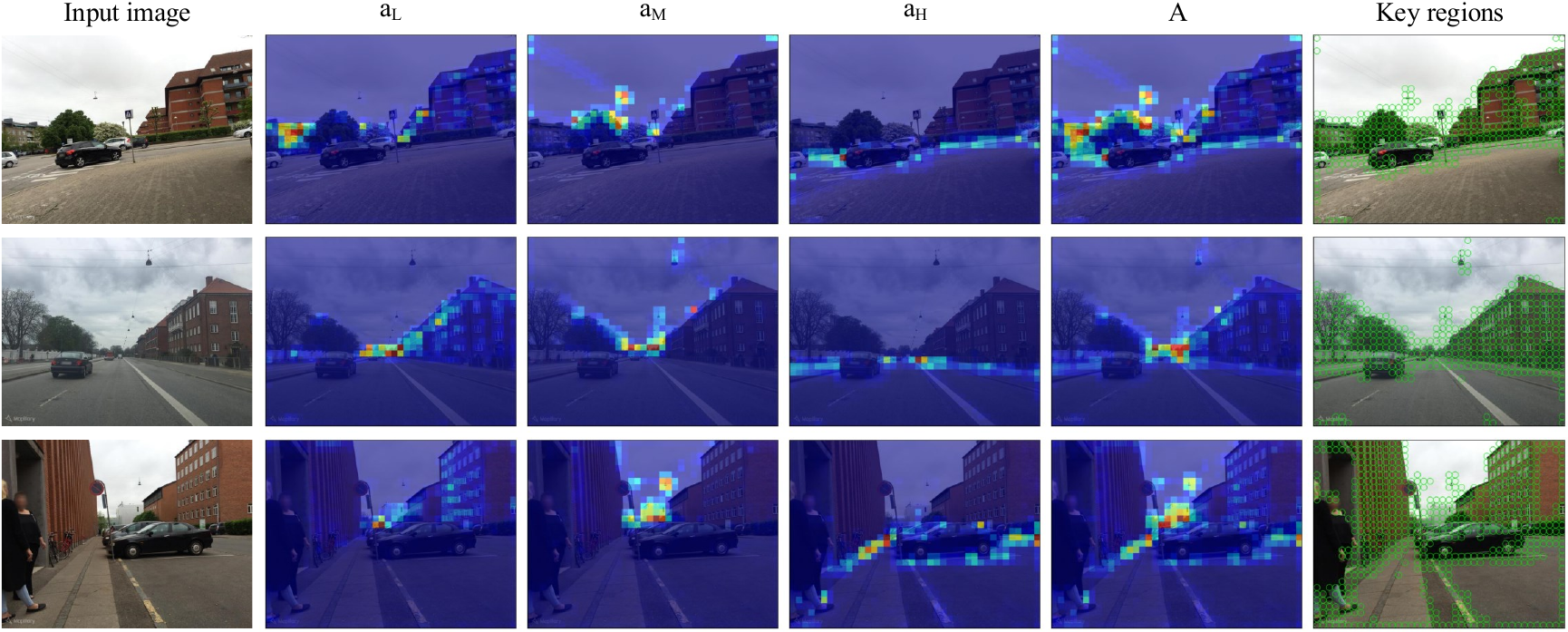}
   \caption{\textbf{Some visualisations on multi-level attentions.} From left to right: Input images, attention maps from each level, final attention maps, output key-patches. All images used here are from MSLS validation set and unseen during the network training stage. Different semantic cues are captured by attention maps from different levels. The network only pays attention to distinctive regions in the image and filters out confusing information. }
   \label{fig:vis_attetion}
\end{figure*}

\subsection{Qualitative Results}
\cref{fig:vis_compare} illustrates some retrieval and matching results of hard examples with challenging conditions. In these cases, TransVPR produces correct matches while all other methods fail. For example, observing the first and fourth rows, where there are severe viewpoint changes or occlusions cased by dynamic objects, TransVPR can successfully perform matching based on distinctive regions and avoid distracting areas. However, other methods show a tendency to retrieve images with similar global layout as the query.

To further have an intuitive interpretation of the semantic cues captured by multi-level attentions, some visualization examples of learned attention maps are presented in \cref{fig:vis_attetion}. This confirms attention maps from different levels tend to focus on areas with different semantic information. For example, $\mathbf{a_L}$ mainly focuses on the small objects and textural areas on the surface of buildings. $\mathbf{a_M}$ focuses on objects in the air, such as street lamps and tree canopies, while $\mathbf{a_H}$ outlines the contours of the ground and the lane lines. All these attention maps avoid distracting areas such as the sky, the ground, dynamic objects, and untextured walls, which may change over time or have no effect on recognizing a scene. Note that we did not add any semantic constraints during the training process. These semantic information can be learned automatically by the attention mechanism in TransVPR under only image-level supervision. 

\subsection{Latency and Memory}

In real-world VPR systems, latency and scalability are important factors that need to be considered. \cref{tab:latency_memory} shows the computational time and memory requirements for all compared techniques to process a single query image. TransVPR is 4.4 times and 29.7 times faster than DELG and Patch-NetVLAD-p in terms of feature encoding, when 11.3 times and 2.4 times faster than them in terms of spatial matching. 

The memory footprint of TransVPR is 1.17 MB per image, the same order as SP-SuperGlue and Patch-NetVLAD-s. Considering patch features account for the main part of storage, using sparse and low-dimensional patch features can reduce the memory cost substantially. Patch-NetVLAD-p has an extremely large memory footprint due to its multi-scale features and high dimensionality (dim = 4096), while TransVPR requires less memory with relatively low-dimensional patch features (dim = 256). 

\begin{table*}
  \centering
  \scalebox{0.75}{
  \renewcommand{\arraystretch}{1.2}
  \begin{tabular}{@{}cl||ccc||ccc||ccc||ccc@{}}
  \toprule
\multicolumn{2}{c||}{\multirow{2}{*}{Method}}  & \multicolumn{3}{c||}{MSLS val}& \multicolumn{3}{c||}{Nordland test} & \multicolumn{3}{c||}{Pitts30k test} & \multicolumn{3}{c}{Robotcar-S2 test}\\
\cline{3-14}
&& R@1 & R@5 & R@10 & R@1 & R@5 & R@10  & R@1 & R@5 & R@10  & .25m/2° & .5m/5° & 5.0m/10° \\
\hline
\multirow{4}{*}{\begin{tabular}[c]{@{}c@{}}Global\\ retrieval\end{tabular}} & 
sL-sATT & 69.2 & 84.6 & 88.9 & 13.1 & 33.9 & 45.5 & 68.6 & 85.2 & 90.3 & 2.2 & 10.6 & 55.6\\
& mL-sATT & 70.7 & 84.3 & 88.0 & 13.7 & 34.7 & 45.6 & 71.2 & 86.5 & 90.8 & 2.5 & 11.0 & 54.9\\
& mL-mATT-plain & \textbf{71.5} & \textbf{85.7} & \textbf{89.9} & 13.2 & 32.7 & 43.0 & 71.2 & 87.0 & 91.3 & \textbf{3.3} & \textbf{12.0} & 56.2\\
& \textbf{mL-mATT-standard} & 70.8 & 85.1 & 89.6 &  \textbf{15.9} & \textbf{38.6} & \textbf{49.4} & \textbf{73.8} & \textbf{88.1} & \textbf{91.9} &2.9 & 11.4 & \textbf{58.6}\\
\hline
\multirow{4}{*}{Re-ranking} & 
sL-sATT & 87.4 & 92.7 & 93.2 &  54.0 & 70.1 & 73.7 & 87.4 & 94.0 & 95.3 & 9.4 & 32.3 & 78.1\\
& mL-sATT & \textbf{87.7} & \textbf{91.5} & \textbf{93.0} & 55.7 & 70.8 & 74.2 & 88.6 & 94.7 & 96.0 & 9.4 & 33.0 & 77.3\\
& mL-mATT-plain & 84.7 & 89.6 & 91.5 & 54.1 & 68.4 & 71.7 & 88.1 & 94.3 & 95.5 & 9.5 & 34.2 & 78.3\\
& \textbf{mL-mATT-standard} & 86.8 & 91.2 & 92.4 &  \textbf{58.8} & \textbf{75.0} & \textbf{78.7} & \textbf{89.0} & \textbf{94.9} & \textbf{96.2} &\textbf{9.8} & \textbf{34.7} & \textbf{80.0}\\
\bottomrule
\end{tabular}}
  \caption{\textbf{Ablations on multi-level attention aggregation strategy.} The proposed TransVPR configuration (mL-mATT-standard) achieves the best results.}
  \label{tab:ablation}
\end{table*}

\begin{table}
  \centering
  \scalebox{0.75}{
 \renewcommand{\arraystretch}{1.2}
\begin{tabular}{c|ccc|ccc}
\toprule
\multirow{2}{*}{\begin{tabular}[c]{@{}c@{}}Attention \\ mask\end{tabular}} & \multicolumn{3}{c|}{MSLS val} & \multicolumn{3}{c}{Nordland test} \\
\cline{2-7}
& R@1      & R@5     & R@10    & R@1       & R@5       & R@10      \\
\hline
None & 81.2     & 87.6    & 90.1    & 58.8      & 73.9      & 77.7      \\
$\mathbf{a_L}$ & 86.4     & 91.1    & 92.2    & 55.1      & 72.9      & 76.9      \\
$\mathbf{a_M}$ & 84.1     & 90.4    & 91.8    & 47.8      & 69.1      & 74.3      \\
$\mathbf{a_H}$ & 61.2     & 77.3    & 82.8    & 23.5      & 40.7      & 48.6      \\
$\mathbf{a_L}$ \& $\mathbf{a_M}$ & \textbf{86.9}     & 90.9    & 91.9    & 57.4      & 74.3      & 77.7      \\
\hline
$\mathbf{A}$ & 86.8 & \textbf{91.2} & \textbf{92.4} & \textbf{58.8} & \textbf{75.0} & \textbf{78.7} \\
\bottomrule
\end{tabular}}
\caption{Performance of TransVPR when using different combinations of attention masks learned from multiple Transformer levels to select key-patch descriptors. The fused multi-level attention mask $\mathbf{A}$ performs best.}
\label{tab:attention_mask_abl}
\end{table}

\subsection{Ablations and Analysis}
We conduct several ablation experiments to further validate the design of TransVPR.

\begin{figure}[t]
  \centering
  \subfloat[MSLS val]{
    \label{fig:nordland}
    \includegraphics[width=.4\linewidth]{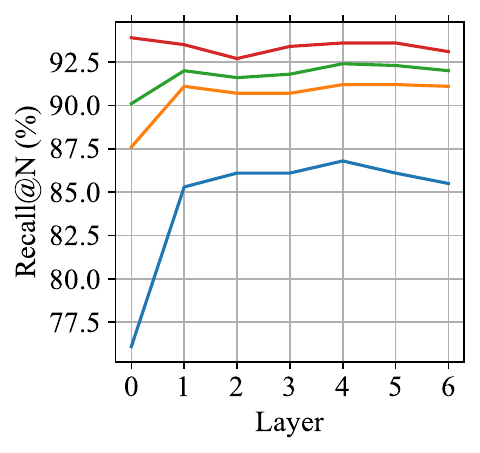}}
  \subfloat[Nordland]{
    \label{fig:msls_val}
    \includegraphics[width=.58\linewidth]{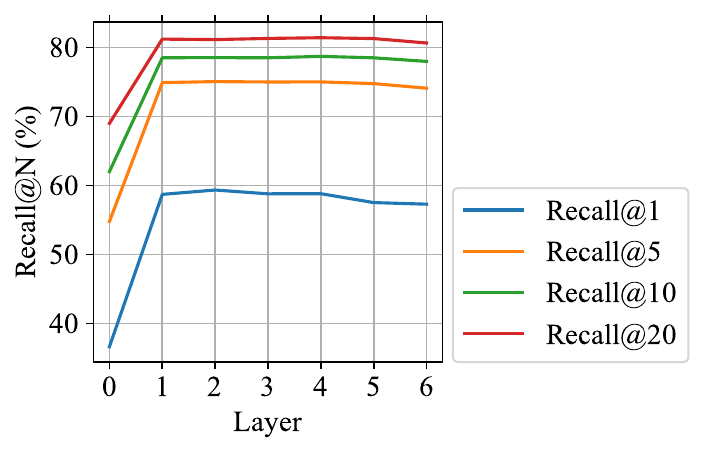}}
  \caption{\textbf{Ablations of local descriptor set selection.} Recall performance of TransVPR with local descriptors from varying Transformer layer. Patch descriptors from any Transformer layer have similar performances and significantly outperform raw patch descriptors (\ie, layer 0). Besides, a slight degradation of performance is observed at the last layer.}
  \label{fig:choice_local}
\end{figure}

\textbf{Choice of Patch Descriptor Sets. }
In \cref{fig:choice_local}, we show the performance of TransVPR when using different patch descriptor sets. Patch descriptors outputted from Transformer layers significantly outperform raw patch descriptors, demonstrating that the global contextual information encoded in Transformer tokens can improve the patch representation. The performances of patch tokens are similar no matter which Transformer layer they are from. Because despite there is no patch-level supervision while training, the patch tokens retain some locality which ensures the accuracy of spatial matching. It might because the residual connections in Transformers give the output tokens the ability to retain original information. However, there is a slight decay of performance when using patch tokens from the last layer which may contain relatively more contextual information.

\textbf{Multi-Level Attentions for Key-Patch Detection.}
To verify the effectiveness of key-patch detection using the fused multi-level attention mask $\mathbf{A}$, we evaluated the TransVPR performance using patch descriptors ($\mathbf{P_M}$) without filtering and with filtering by each individual attention mask or by their combinations. The results are shown in \cref{tab:attention_mask_abl}. While using individual attention masks, the low-level attention mask $\mathbf{a_L}$ achieves the best performance, and the performance of $\mathbf{a_H}$ is 
dramatically poor. It indicates that building surfaces and fixed objects contribute the most to place recognition. Attention mask combining $\mathbf{a_L}$ and $\mathbf{a_M}$ achieves better result than using any single attention mask, but its performance is still inferior than $\mathbf{A}$. Besides, matching using all patch descriptors without filtering leads to weak performance compared with $\mathbf{A}$.

\textbf{Multi-Level Attention Aggregation Strategy.}
We study how the proposed multi-level attention aggregation strategy affects the model performance by comparing the standard TransVPR with three degenerate configurations:
\begin{itemize}
    \item \textit{Multi-level \& multiple attention maps \& plain connection} (mL-mATT-plain). We remove the concatenation operation of multi-level patch tokens before computing attention maps. The three attention maps are computed only using patch tokens from the same level.
    \item \textit{Multi-level \& single attention map} (mL-sATT). Instead of estimating three attention maps separately, a single attention map $\mathbf{A}$ is computed based on concatenated patch tokens $\mathbf{P}$. The global feature is expressed by the summation of $\mathbf{P}$ weighted by $\mathbf{A}$.
    \item \textit{Single level \& single attention map} (sL-sATT). Only patch tokens from the last Transformer layer are used to estimate both the attention map and the global feature.
\end{itemize}

Detail architectures of these configurations are illustrated in Supplementary Material. \cref{tab:ablation} presents the evaluation results. There is in general a performance degradation from the standard TransVPR (mL-mATT-standard) to mL-sATT and then to sL-sATT, and mL-mATT-standard largely outperforms mL-mATT-plain on all datasets after re-ranking. Besides, the standard TransVPR has a better generalization ability on datasets with data distribution very different from the training set. These results demonstrate the effectiveness of combining multi-level information across Transformer layers and estimating separate attention maps for each level. In addition, results of all configurations are significantly improved by re-ranking using key-patch descriptors, especially on Nordland dataset which suffers severe perceptual aliasing and thus relies more on fine-grained spatial matching.   

\section{Conclusion}
In this work, we have designed a novel vision Transformer-based place recognition model, TransVPR, which jointly extracts distinctive global and patch-level image features by aggregating multi-level attentions. All components of TransVPR are integrated in a single and lightweight network, enabling end-to-end optimization with image-level supervision. TransVPR outperforms some state-of-the-art VPR techniques on several benchmark datasets and achieves superior trade-off between accuracy and efficiency. Ablation results further verify the effectiveness of the design of our model.

This demonstration of TransVPR is limited in VPR tasks, where the major limitation is that the camera localization would be not precise enough when reference images are sparse. Therefore, one research topic of future work is to estimate camera poses in a regression framework by exploiting the TransVPR descriptors.

\section*{Acknowledgement}
\begin{flushleft}
\hspace{1em} This work was supported by the National Natural Science Foundation of China under grant No. 62088102 and No. 62106192.
\end{flushleft}
{\small
\bibliographystyle{ieee_fullname}
\bibliography{reference}
}

\end{document}